\documentclass[preprint,12pt]{elsarticle}

\usepackage[bookmarks=false]{hyperref}
\usepackage{graphicx}
\usepackage{pifont}
\usepackage{amsmath,amssymb}
\hypersetup{draft}

\begin{document}

\begin{frontmatter}

\title{Detection of Virus and Small Cell Patches in Foci Images Using Switchable Convolution and Feature Pyramid Networks}

\author{Amrita Singh, Snehasis Mukherjee}



\begin{abstract}
Accurate detection and counting of virus patches in focus-forming unit (FFU) images, also known as foci images, are important for quantifying viral infection and analyzing cellular structures. This task is challenging because biomedical targets often vary substantially in size, density, contrast, and shape. In this paper, we propose an enhanced YOLOv2-based detector that integrates a Feature Pyramid Network (FPN) to improve multi-scale feature representation. We also incorporate a switchable atrous convolution mechanism to adapt the receptive field for fine-grained targets in dense microscopy images. The proposed method is evaluated on biomedical foci image datasets for virus patch and small cell patch detection. For small cell patch detection, the model achieves a mean average precision (mAP) of 40.5\% at a 25\% Intersection over Union (IoU) threshold. For FFU virus patch detection, the model achieves an mAP of 68\%. These results indicate that combining FPN-based feature fusion with switchable convolution improves the suitability of YOLOv2 for specialized biomedical object detection tasks.
\end{abstract}


\end{frontmatter}

\section{Introduction}
Small-object detection is a central requirement in many computer vision applications, including autonomous driving, medical diagnosis, microscopy, and forensic analysis. Despite substantial progress in object detection, accurately localizing small targets remains difficult because such objects often occupy only a few pixels, provide limited contextual information, and appear in noisy or cluttered backgrounds. These challenges are particularly important in biomedical microscopy, where small cell patches and virus-induced foci must be identified accurately for downstream analysis and quantification.

In medical imaging, reliable detection of small cell patches is essential for analyzing disease progression and cellular behavior \cite{wang2023small}. However, small targets are easily missed or confused with background artifacts, leading to false negatives and false positives. Several methods have been proposed to address this problem. For example, Pixel Level Balancing (PLB) dynamically adjusts the training loss according to object size to improve small-object precision in medical images \cite{hu2022small}. Multi-scale feature fusion has also been widely adopted, with Feature Pyramid Networks (FPNs) combining features from different network depths to represent objects across a range of scales \cite{zhou2025multi}.

Deep learning has transformed object detection by enabling convolutional neural networks (CNNs) to learn hierarchical representations directly from data. A detailed review of small-object detection strategies, including input-resolution enhancement, scale-aware training, and contextual modelling, is provided in \cite{feng2023deep}. Among common detection frameworks, Faster R-CNN \cite{ren2015faster}, SSD \cite{liu2016ssd}, and YOLO \cite{redmon2016you} have achieved strong performance on general object detection benchmarks. YOLOv2 extends the original YOLO framework by introducing batch normalization, anchor boxes, and multi-scale training \cite{redmon2017yolo9000}. Its favorable balance between speed, simplicity, and detection performance makes it attractive for resource-constrained biomedical applications.

In FFU assays, viral infectivity is quantified by identifying and counting virus-induced foci in microscopy images. Manual analysis is time-consuming and subjective, while conventional image-processing approaches struggle with variations in patch size, shape, contrast, and density. Although YOLOv2 is efficient, its original architecture relies primarily on a single high-level feature map. This design can reduce sensitivity to small objects because fine spatial details are progressively lost through downsampling.

Recent YOLO variants, such as YOLOv8 \cite{li2024improving}, include stronger multi-scale mechanisms through improved backbones, feature pyramids, path aggregation, and anchor-free prediction. However, such models can be unnecessarily complex for small biomedical datasets and may be more prone to overfitting. In this work, we retain the simpler YOLOv2 framework and enhance it with FPN-based multi-scale fusion \cite{lin2017feature}. We further introduce a switchable atrous convolution mechanism, following \cite{singh2024automatic}, to adapt the receptive field for virus patches and small cell patches. The main contributions of this work are as follows.
\begin{itemize}
    \item We integrate an FPN module into YOLOv2 to improve multi-scale feature extraction for small biomedical targets.
    \item We incorporate switchable atrous convolution to adapt the receptive field to virus patches and small cell structures.
\end{itemize}

\section{Related Work}
\label{sec:related_work}
Object detection has advanced rapidly over the past decade, largely due to deep learning-based frameworks \cite{feng2023deep}. Earlier approaches relied on handcrafted features, sliding windows, and models such as Deformable Part Models (DPM) \cite{girshick2015deformable}. Although effective in some settings, these methods were limited by hand-designed representations, high computational cost, and reduced scalability.

Modern detectors, including Faster R-CNN \cite{ren2015faster}, SSD \cite{liu2016ssd}, and YOLO \cite{Redmon_2016_CVPR}, address these limitations by learning features directly from data. YOLO is particularly attractive for real-time inference, whereas SSD and Faster R-CNN often provide strong accuracy in more complex detection scenarios. YOLOv2 \cite{redmon2017yolo9000} further improves the trade-off between speed and accuracy, making it suitable for a wide range of practical applications.

Biomedical object detection introduces additional challenges because target structures vary widely in size, shape, intensity, and appearance. Applications such as cell counting, tumor detection, and virus quantification require both high precision and robustness. Traditional biomedical image-analysis pipelines often use thresholding and morphological operations~\cite{sezgin2004thresholding}; however, these methods can be sensitive to imaging conditions and usually require dataset-specific parameter tuning.

Deep learning-based methods have shown considerable promise in this domain. U-Net~\cite{ronneberger2015u}, for example, is widely used for biomedical segmentation tasks such as cell and organ delineation. Faster R-CNN and YOLO have also been adapted for biomedical detection tasks~\cite{Shen2017}, including polyp detection, blood cell identification, and lesion detection. Nevertheless, detecting very small and irregularly shaped objects remains challenging.

Feature Pyramid Networks (FPNs)~\cite{lin2017feature} are an important enhancement for both general and biomedical object detection. By combining high-level semantic features with lower-level spatial features, FPNs improve the representation of small and densely packed objects. This study integrates FPN-based feature fusion into YOLOv2 and combines it with switchable atrous convolution to address the challenges of detecting virus patches and small cell structures with varying shapes and sizes in FFU images.

\section{Methodology}\label{sec:methodology}
The proposed model combines a Feature Pyramid Network (FPN) with a Switchable Atrous Convolution (SAC) mechanism to improve multi-scale feature representation and adapt to objects with varying sizes and shapes. This section first describes the modified YOLOv2 baseline and then presents the SAC and FPN components introduced in the proposed architecture.

\subsection{YOLOv2 Baseline Model}
\begin{figure}[!ht]
     \centering
     \includegraphics[width=\textwidth]{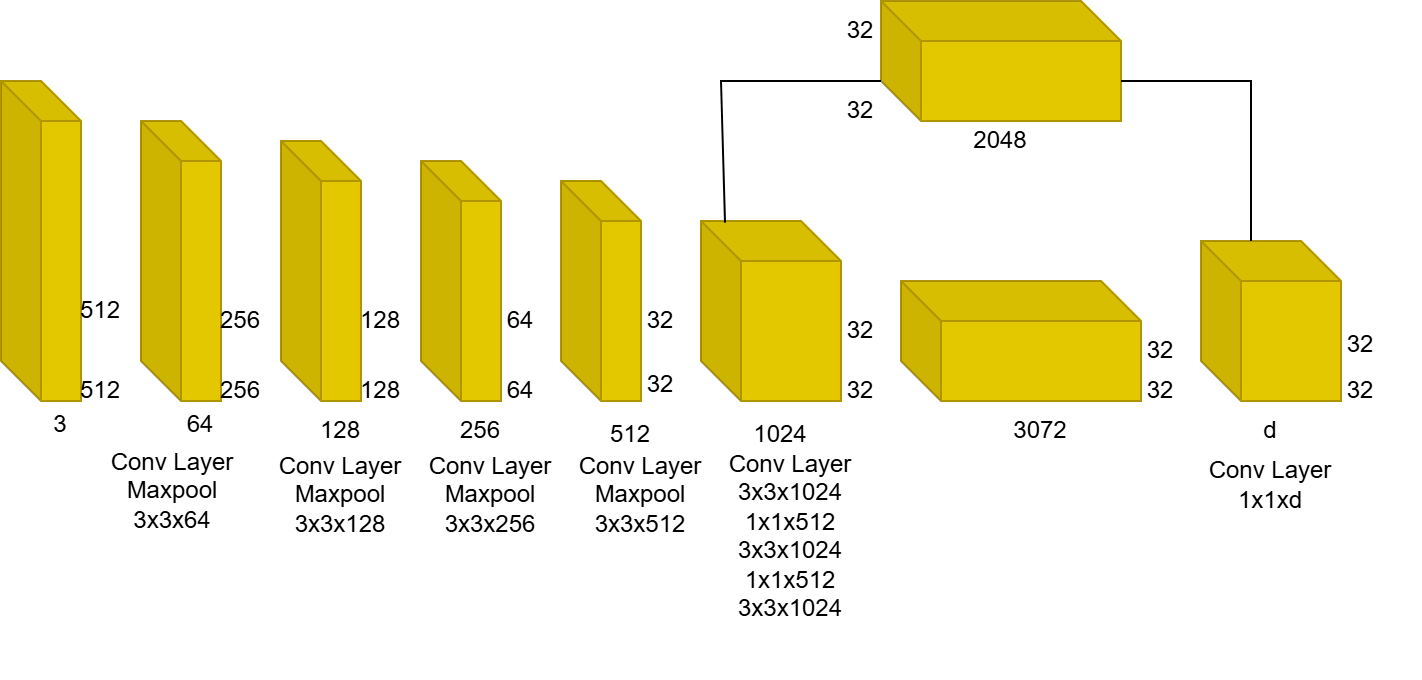}
\caption{Modified YOLOv2 architecture. One max-pooling layer is removed from Module 6, following \cite{singh}.}
\label{fig:2}
 \end{figure}

YOLOv2 is a single-stage detector that predicts bounding boxes and class probabilities directly from convolutional feature maps \cite{redmon2017yolo9000}. Its use of high-level semantic features supports efficient inference, but the loss of fine spatial detail can reduce performance for small objects. YOLOv2 uses Darknet-19 as its backbone, a lightweight CNN composed of 19 convolutional layers, five max-pooling layers, Leaky ReLU activations, batch normalization, and mainly $3 \times 3$ and $1 \times 1$ kernels. To improve spatial resolution, we modify the Darknet-19 backbone by removing one max-pooling layer. We also increase the input resolution from $416 \times 416$ to $512 \times 512$, increasing the output grid from $13 \times 13$ to $32 \times 32$. Figure~\ref{fig:2} shows the modified YOLOv2 architecture.

\subsection{Switchable Atrous Convolution (SAC)}
\begin{figure}[!ht]
     \centering
     \includegraphics[width=\textwidth]{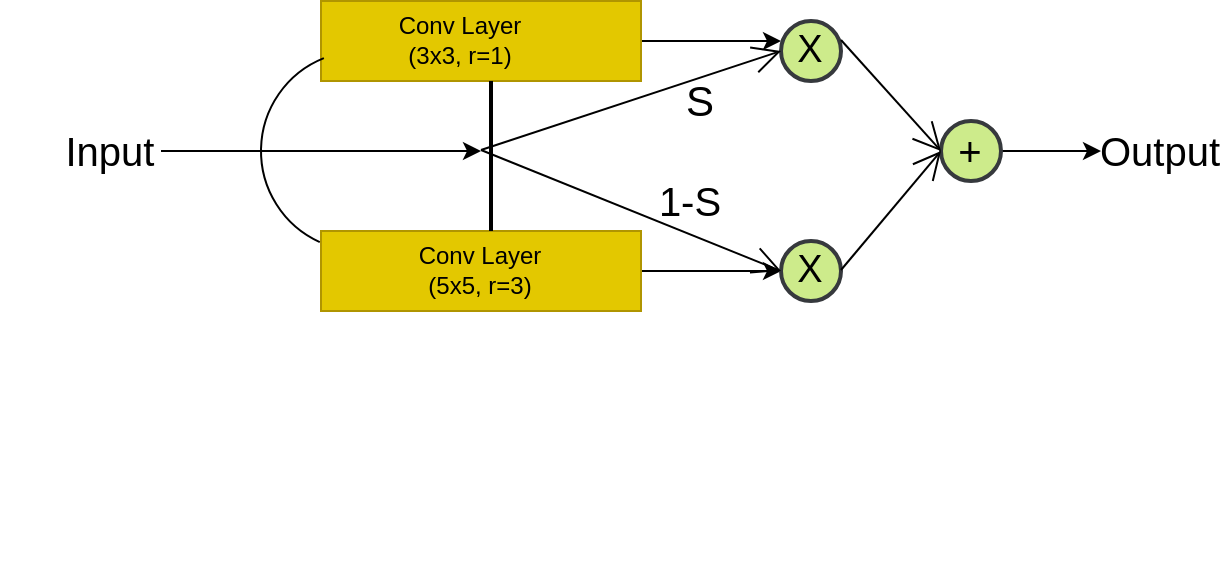}
\caption{The Switchable Atrous Convolution (SAC) block used in the proposed model.}
     \label{fig:3}
 \end{figure}

To adapt dynamically to variations in object size and structure, we integrate a Switchable Atrous Convolution module following \cite{singh}. The SAC block uses parallel convolutional paths with different receptive fields and learns to combine their feature responses according to the input. In the modified YOLOv2 model, feature maps at resolutions of $512 \times 512$, $256 \times 256$, $128 \times 128$, and $64 \times 64$ are passed through switchable convolutional layers, as illustrated in Figure~\ref{fig:3}. These layers use parallel atrous convolutions with $3 \times 3$ and $5 \times 5$ kernels. The resulting feature maps are then forwarded to the corresponding bottom-up layers of the FPN.

\subsection{Feature Pyramid Network (FPN)}
\begin{figure}[!ht]
    \centering
    \includegraphics[width=\textwidth]{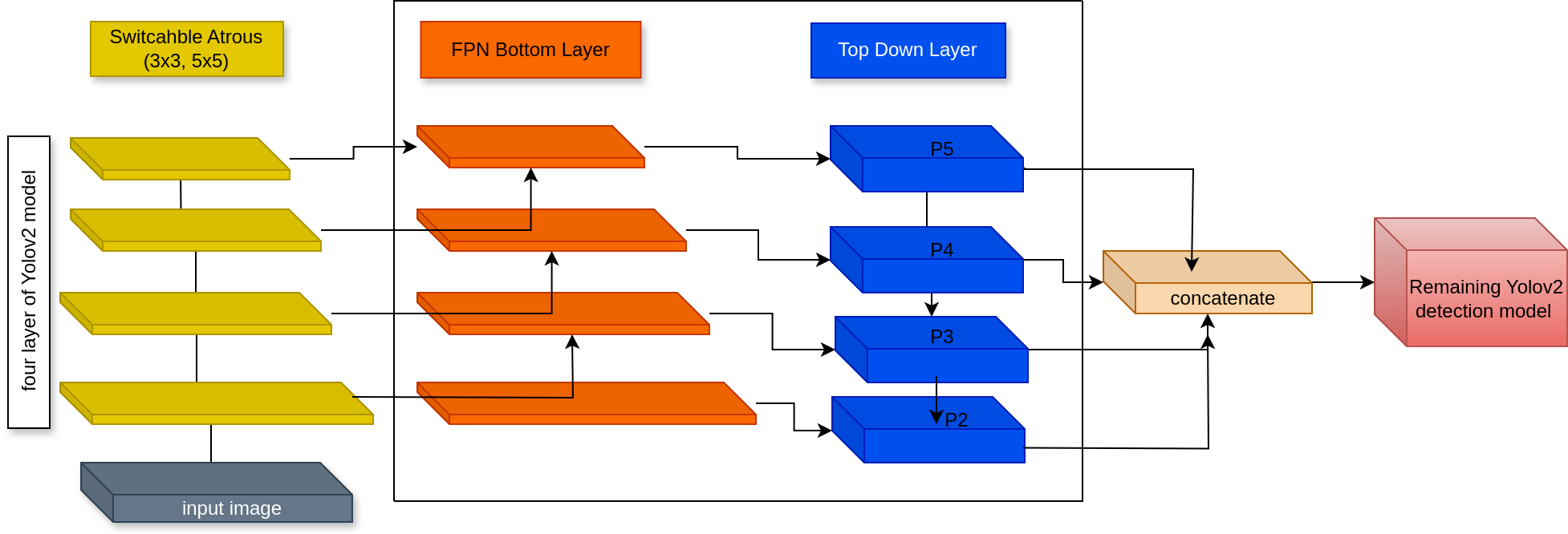}
    \caption{Overall architecture of the proposed model for small patch detection.}
    \label{fig:4}
\end{figure}

To address YOLOv2's limitations in small-object detection, we incorporate an FPN, as shown in Figure~\ref{fig:4}. The FPN improves multi-scale representation by extracting hierarchical features from different backbone depths, upsampling high-level semantic features, and merging them with lower-level spatial features through lateral connections. It constructs feature maps at multiple scales (P2, P3, P4, and P5), which are further processed and concatenated to form a richer representation. This design helps the detector capture both fine local details and broader contextual information. Figure~\ref{fig:4} shows the overall proposed architecture.

\section{Results}
\label{sec:experiments}
The proposed method was evaluated on both the virus patch dataset and the small cell patch dataset \cite{singh}. We report quantitative detection performance and qualitative examples, and we discuss the contribution of the SAC and FPN components through ablation comparisons.

\subsection{Quantitative Results}
Table~\ref{tab1} summarizes the mAP values obtained by the compared models. On the virus patch dataset, SAC\_YOLOv2 achieved the highest reported mAP of 76.19\%, outperforming the newer YOLO variants included in the comparison. This result suggests that, for the available dataset size, a simpler YOLOv2-based architecture enhanced with SAC and FPN can be more effective than deeper models. SAC\_YOLOv7 improved over the baseline YOLOv7, reaching an mAP of 56\%, but it did not match the performance of the YOLOv2-based variants, likely because of the higher complexity of the model.
\begin{table}[htbp]
\caption{Comparison of detection performance for FFU virus patches and small cell patches.}
\begin{center}
\resizebox{\textwidth}{!}{%
\begin{tabular}{|l|l|l|l|l|l|}
\hline
\textbf{Model Name}&\multicolumn{4}{|l|}{\textbf{FFU virus patch detection}} \\
\cline{1-6} 
\textbf{} & \textbf{\textit{Epoch}}& \textbf{\textit{Learning Rate}}& \textbf{\textit{Optimizer}} &\textbf{\textit{Batch Size}}&\textbf{\textit{ mAP}}  \\
\textbf{YOLOv2\_FPN\_Switch} & 100& 0.00001&Adam & 8& \textbf{68.81} \\
 YOLOv2  & 100 & .00001 & Adam & 4 & 65.23 \\

YOLOv7  & 500 & .00001  & Adam  & 4 & 50 \\
SAC\_YOLOv2\cite{singh2024automatic}  & 100 & .00001  & Adam  & 4  & 76.19 \\

SAC\_YOLOv7 \cite{singh2024automatic} & 500 & .00001  & Adam  & 4 & 56 \\
\hline
\textbf{Model Name}&\multicolumn{4}{|l|}{\textbf{Small cell patch detection}} \\
\cline{1-6} 
\textbf{} & \textbf{\textit{Epoch}}& \textbf{\textit{Learning Rate}}& \textbf{\textit{Optimizer}} &\textbf{\textit{Batch Size}}&\textbf{\textit{ mAP}}  \\
\textbf{YOLOv2\_FPN\_Switch} & 150 & 0.00001 & Adam & 8 & \textbf{40.55 }\\
YOLOv2\_FPN  & 200 &  0.001 & Adam  & 4 & 39.28 \\

YOLOv7 & 500 & .00001 & Adam & 4 & 12 \\

SAC\_YOLOv2  & 100 & .00001  & Adam  & 4 & 24.34 \\
SAC\_YOLOv7 & 500 & .00001  & Adam  & 4 & 13 \\


\hline
\end{tabular}%
}
\label{tab1}
\end{center}
\end{table}

On the small cell patch dataset, the proposed YOLOv2\_FPN\_Switch model achieved the best mAP of 40.55\%, followed by YOLOv2\_FPN with 39.28\%. SAC\_YOLOv2 achieved 24.34\%, indicating that switchable convolution alone is insufficient for this dataset, whereas combining switchable convolution with FPN-based feature fusion improves detection of small and variable targets.

\subsection{Ablation Studies}
\begin{figure}[!ht]
    \centering
    \includegraphics[width=0.5\textwidth]{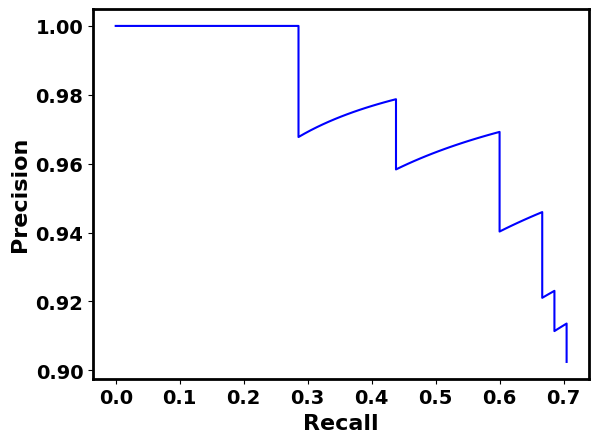}
    \caption{Precision--recall behavior of YOLOv2\_FPN\_SAC for virus patch detection.}
    \label{fig:5}
\end{figure}

The ablation results in Table~\ref{tab1} indicate that FPN-based feature fusion improves detection by leveraging multi-scale information. The SAC mechanism further adapts feature extraction to targets with different apparent sizes and shapes, making the combined model suitable for challenging biomedical microscopy images.
Figure~\ref{fig:5} presents the precision--recall (PR) behavior for virus patch detection. PR curves are particularly useful for imbalanced detection tasks because they show the trade-off between detecting more true positives and maintaining high prediction confidence. The curve begins at high precision for the most confident detections and decreases as recall increases, reflecting the inclusion of more difficult predictions at lower confidence thresholds.

\begin{figure}[!ht]
    \centering
    \includegraphics[width=\textwidth]{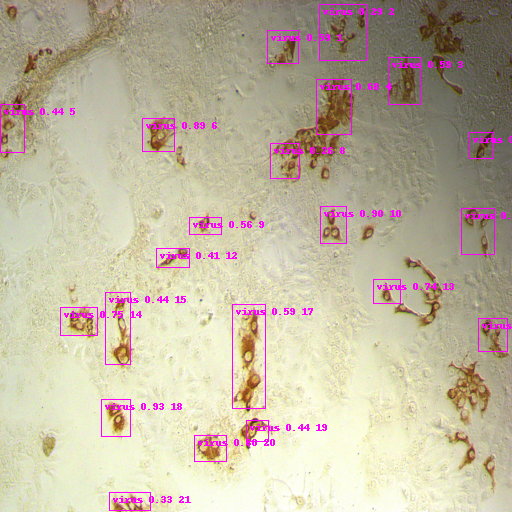}
    \caption{Qualitative virus patch detection results obtained using the proposed YOLOv2\_FPN\_SAC model.}
    \label{fig:6}
\end{figure}

    

Figure~\ref{fig:7} shows the precision--recall curve for small cell patch detection. The maximum recall is approximately 0.67, indicating that the model detects about 67\% of the ground-truth objects at the evaluated thresholds. The gradual decrease in precision illustrates the difficulty of maintaining high detection quality for small or ambiguous targets. Overall, the PR curve indicates reliable performance for high-confidence predictions, while additional feature-refinement strategies may further improve recall without substantially reducing precision.
\begin{figure}[!ht]
   \centering
 \caption{Precision–Recall curve for the YOLOv2-FPN-Switchable Atrous Convolution model.}

  \label{fig:7}
    
\end{figure}

\subsection{Qualitative Analysis}
Figure~\ref{fig:6} illustrates representative detection outputs. The qualitative results show that the SAC-enhanced model can identify small and dense structures in microscopy images. Each detected cell is marked with a magenta bounding box and a label in the following format:

\begin{center}
\texttt{Cell: \{Confidence Score\} \{Detection Index\}}
\end{center}

\textbf{Confidence Score:} the model's confidence in the prediction. Values closer to 1 indicate higher certainty that the detected region corresponds to a virus-infected cell.

\textbf{Detection Index:} a unique identifier assigned to each detected instance for referencing or counting.

The results demonstrate the ability of the model to detect both isolated and clustered infected cells across a range of scales. The SAC module supports this behavior by adapting the effective receptive field of convolutional filters, allowing the network to capture both fine-grained details and contextual cues.

The detected patches include early, marginal, and widespread infection patterns. The high detection density in clustered regions indicates that the model is sensitive to densely packed viral formations. These qualitative results support the use of the proposed detector for automated infection quantification in biomedical image analysis, particularly when targets are small, low contrast, or irregularly shaped.

    

    

\section{Conclusion}
\label{sec:conclusion}
This paper presented a YOLOv2-based framework for detecting virus patches and small cell patches in foci images. The method enhances the baseline detector with two complementary components: a switchable atrous convolution block that adapts the receptive field to objects of varying size, and an FPN module that fuses multi-scale features for improved localization of small and irregular targets. The experimental results show that the proposed configuration improves detection performance on challenging biomedical microscopy datasets while retaining the simplicity of the YOLOv2 backbone. Future work will explore stronger feature-refinement modules, broader validation on additional biomedical datasets, and extensions of the FPN-enhanced design to other biomedical object detection tasks.
\bibliographystyle{elsarticle-num}
\bibliography{ref}

\end{document}